%% file: main.tex
\pdfoutput=1

\documentclass{article}

\usepackage{PRIMEarxiv}
\usepackage{natbib}

\usepackage{booktabs}

\usepackage[utf8]{inputenc} 
\usepackage[T1]{fontenc}    
\usepackage{hyperref}       
\usepackage{url}            
\usepackage{amsfonts}       
\usepackage{nicefrac}       
\usepackage{microtype}      
\usepackage{amsmath,amssymb,latexsym}
\usepackage[colorinlistoftodos]{todonotes}
\input{macros}

\usepackage{float}
\usepackage{subcaption}

\title{Mortality Prediction Models with Clinical Notes Using Sparse Attention at the Word and Sentence Levels
\thanks{\textit{\underline{Citation}}: 
\textbf{Technical Reports at the Department of Medical Informatics, Amsterdam UMC, 2021. \url{https://kik.amc.nl/KIK/reports/TR2021-01.pdf}}} 
}

\author{
 
  Miguel Rios\\
  Centre for Translation Studies \\
  University of Vienna \\
  \texttt{miguel.angel.rios.gaona@univie.ac.at} \\

   \And
  Ameen Abu-Hanna  \\
  Department of Medical Informatics, Amsterdam UMC \\
  University of Amsterdam \\
  \texttt{a.abu-hanna@amsterdamumc.nl} \\
}

\begin{document}

\maketitle

\begin{abstract}
  Intensive Care in-hospital mortality prediction has various clinical applications. Neural prediction models, especially when capitalising on clinical notes, have been put forward as improvement on currently existing models. However, to be acceptable these models should be performant and transparent. This work studies different attention mechanisms for clinical neural prediction models in terms of their discrimination and calibration. Specifically, we investigate sparse attention as an alternative to dense attention weights in the task of in-hospital mortality prediction from clinical notes. We evaluate the attention mechanisms based on: i) local self-attention over words in a sentence, and ii) global self-attention with a transformer architecture across sentences. We demonstrate that the sparse mechanism approach outperforms the dense one for the local self-attention in terms of predictive performance with a publicly available dataset, and puts higher attention to prespecified relevant directive words. The performance at the sentence level, however, deteriorates as sentences including the influential directive words tend to be dropped all together.

\end{abstract}

\keywords{sparse attention \and clinical notes \and mortality prediction}

\section{Introduction}
Deep learning has become a promising approach for clinical prediction models \citep{rajkomar2018, Shickel2019DeepSOFAAC}. Moreover, natural language processing (NLP) applications based on neural network (NN) models have shown the potential to benefit the task of mortality prediction \citep{kemp2019improved}.  However, NN models based on clinical notes are difficult to interpret, due to their black-box nature. Decision makers need transparency in the way words in an input clinical note contribute to the overall prediction. One approach to gain transparency is the attention mechanism, which have shown improvements across NLP applications \citep{Vaswani2017, devlin-etal-2019-bert} for text representation. The attention mechanism assigns weights to an input vector representation of a layer in a NN \citep{bahdanau2014neural}, and it has been used to identify the relative importance of a word  \citep{clark-etal-2019-bert, huang2019clinicalbert,vashishth2019attention}.

\citet{caicedo2020} propose a mortality prediction model with clinical notes based on a convolutional architecture, where the max pooling layer is used to inform word importance. \citet{lovelace2019explainable} incorporate an attention mechanism on top of the convolutional representation model to highlight words from the input clinical notes. 
 

Current attention mechanisms produce a dense weight distribution over the input vector representation. The dense attention will assign a weight to all words in the context even if they are irrelevant for the current prediction. Moreover, attention weights may not be interpretable given that their alterations do not lead to a change in predictions \citep{serrano-smith-2019-attention, jain-wallace-2019-attention}. In contrast, sparse attention can lead to more accurate models and more transparent presentation  by assigning zero weights to some words \citep{martins16, niculae2017regularized, correia-etal-2019-adaptively}.

In this paper, we investigate the predictive performance of mortality prediction models, in terms of discrimination and calibration, using clinical notes. In particular, we study the effect of sparse versus dense attention mechanisms within a hierarchical architecture: self-attention is applied at the local level to weigh words in a sentence, and at the global level to weigh sentences. We show preliminary results for the task of in-hospital mortality prediction using a publicly available dataset. First, we compare the predictive performance between a model based on local self-attention over words using the standard dense attention, and two sparse attention mechanisms $\sparsemax$ and $\entmax$. The sparse attention mechanisms show better performance compared to dense attention. A preliminary evaluation showed that $\sparsemax$ consistently identified prespecified directive words known to be associated with mortality. Finally, we use the different attention mechanisms on a global self-attention model across sentences on clinical notes. In contrast to the previous experiment, the global attention model with the dense mechanism outperforms the sparse ones. The latter seems to throw sentences out with the directive words.

\section{Sparse Attention}

The Self-attention or scaled dot-product attention is a main component in current state-of-the-art NLP models \citep{Vaswani2017}.
Self-attention computes the representation of a sequence by processing each of its positions and associating the position with other positions for indications of significance. The self-attention layer is defined as follows:

\begin{subequations}
\label{eq:att}
\begin{align}
    \Att(\boldsymbol{Q}, \boldsymbol{K}, \boldsymbol{V})=\boldsymbol{\pi}\left(\frac{\boldsymbol{Q} \boldsymbol{K}^{\top}}{\sqrt{d}}\right) \boldsymbol{V}
\end{align}
\end{subequations}
where $\boldsymbol{Q}$ indicates queries, $\boldsymbol{K}$ keys, $\boldsymbol{V}$ values, and $d$ vector dimensionality.
The queries are a linear transformation of the input vector $x$ with a parameter matrix $W_q$ defined as $\boldsymbol{Q} = W_q x$, and similarly for keys $\boldsymbol{K} = W_k x$, and values $\boldsymbol{V} = W_v x$.
The standard $\boldsymbol{\pi}$ mapping is based on $\softmax$, and it is used to normalise the attention weights. Moreover, self-attention can be extended to multiple heads that potentially allows the model to focus on different positions \citep{Vaswani2017}.

\citet{martins16} propose $\sparsemax$ as a differentiable mapping that provides a sparse alternative to $\softmax$. Sparsemax projects the input vector into the probability simplex, where it is likely to be in the boundary of the simplex and then becoming sparse.
Moreover, the $\entmax$ mapping produces sparse distributions by defining interpolations between $\softmax$ and $\sparsemax$ \citep{peters-etal-2019-sparse, correia-etal-2019-adaptively}. The $\entmax$ becomes an intermediate mapping between $\sparsemax$ and $\softmax$.
\citet{correia-etal-2019-adaptively} propose a sparse attention mechanism for transformer architectures by replacing the $\pi$ mapping by $\sparsemax$, and $\entmax$. 

Most mortality prediction models based on free text represent multiple notes of an ICU stay as a sequence of sentences. The model hierarchically composes word representations (i.e. word encoder) into a sentence, and sentences (i.e. sentence encoder) into a patient representation  for predicting mortality \citep{grnarovaSHE16, si2019deep}. For example, \citet{grnarovaSHE16} use a hierarchy of convolutional neural networks (CNN) on the word and sentence encoders for mortality prediction given complete ICU stays.

We use a word encoder with one layer self-attention and one head as a baseline to represent local dependencies over words in a sentence. The clinical notes of a patient for an ICU stay $x_{i,t}$ consist of $i$ words and $t$ sentences. The input $x_{i,t}$ is first represented with pre-trained word embeddings, and then projected into a linear layer followed by self-attention (Eq. \ref{eq:att}) to encode words of each sentence. Finally, the features of the word encoder are summarised by averaging to produce a patient representation for a linear prediction layer. 
For the local self-attention, we define the following models: \textbf{Att-softmax} as baseline with $\softmax$ mapping for dense attention weights,  \textbf{Att-sparsemax} with the attention weights based on the $\sparsemax$ mapping, and \textbf{Att-entmax} with the $\entmax$ mapping.

A limitation of a local architecture with self-attention is the lack of long distance dependencies across sentence representations. We use a hierarchical architecture for modelling long distance relations within an ICU stay. We replace the CNN with a transformer layer \citep{Vaswani2017}, which is based on self-attention and positional embeddings, on both levels of the hierarchy \citep{trHAN, zhang-etal-2019-hibert}. For the global attention, we define the following hierarchical transformers: \textbf{Tr-softmax}, \textbf{Tr-entmax}, and \textbf{Tr-sparsemax}. We describe the self-attention and hierarchical transformer architectures in more detail in Appendix \ref{sect:apparch}.

\begin{table*}[ht]
\small
\centering
\begin{tabular}{lccc}
\toprule

Model      & \multicolumn{1}{c}{AUC-ROC$\uparrow$} & \multicolumn{1}{c}{AUC-PR$\uparrow$} &  \multicolumn{1}{c}{Brier$\downarrow$}\\ \hline

Att-softmax & 0.824	$\pm$ 0.003 &      0.435 $\pm$ 0.008    &  0.085 $\pm$	0.001  \\

Att-entmax &  0.827 $\pm$	0.004 &   0.445 $\pm$	0.010        &  0.084	 $\pm$ 0.002\\ 

Att-sparsemax &  0.834 $\pm$ 0.004 &   0.467 $\pm$ 0.011  &  0.087 $\pm$  0.001    \\

\hline
		


		
Tr-softmax & 0.839 $\pm$ 0.005  &      0.462 $\pm$	0.005     &  0.093 $\pm$	0.016\\

Tr-entmax & 	0.828 $\pm$	0.007  &     0.433 $\pm$	0.014       &  0.114 $\pm$ 0.004	\\

Tr-sparsemax &  0.801 $\pm$	0.009 &    0.373 $\pm$	0.025        &  0.196 $\pm$	0.076\\
\bottomrule
\end{tabular}

\caption{In-hospital mortality results on the test dataset with 5 runs $\pm$ standard deviation on each model reporting  AUC-ROC, AUC-PR, and Brier score.}
\label{Tab:inhospital}
\end{table*}

\label{sect:perfig}
\begin{figure*}[h]%
    \centering
     \subfloat[]{{\includegraphics[width=7.5cm]{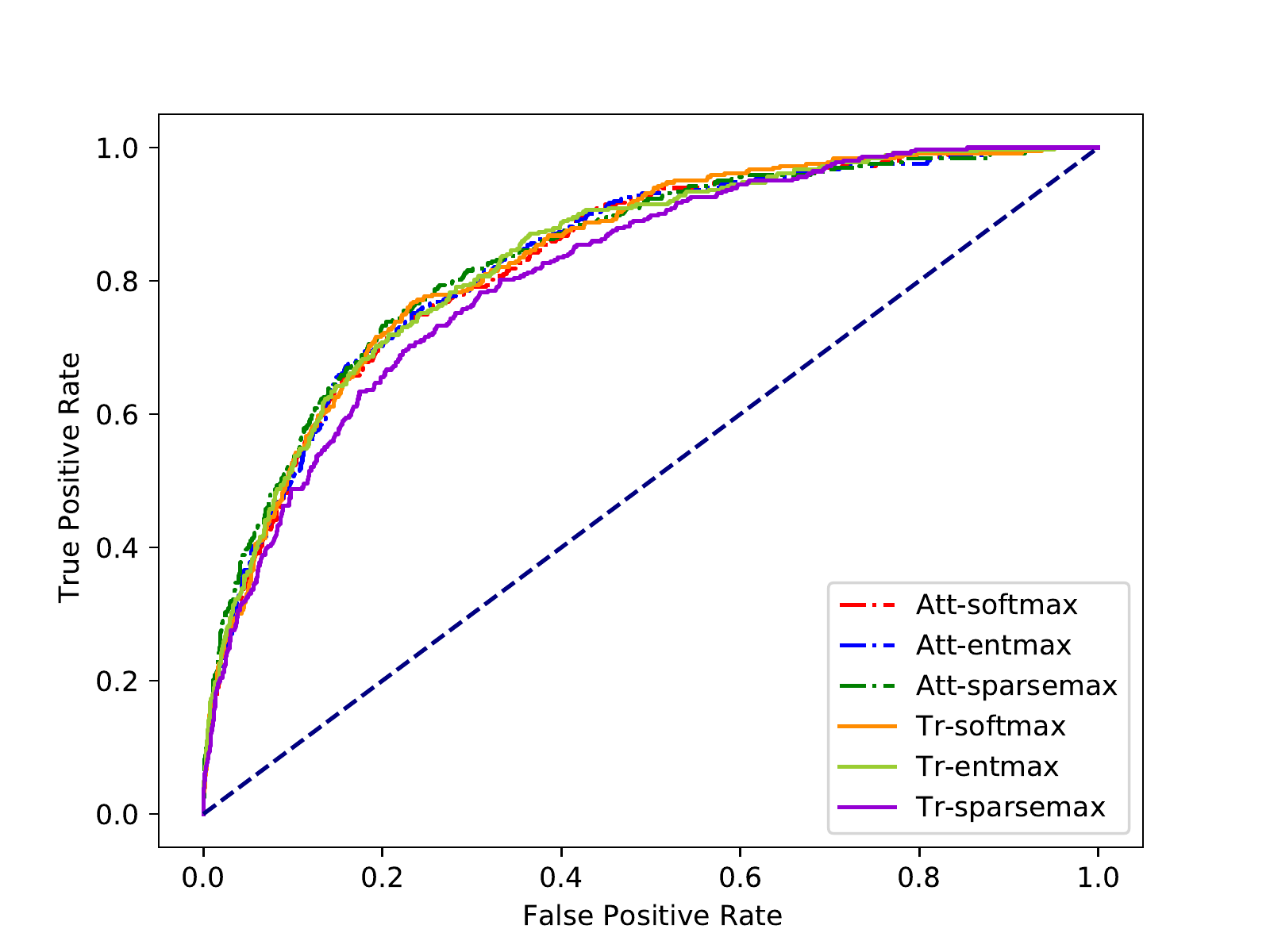} }}%
    \qquad
    \subfloat[]{{\includegraphics[width=7.5cm]{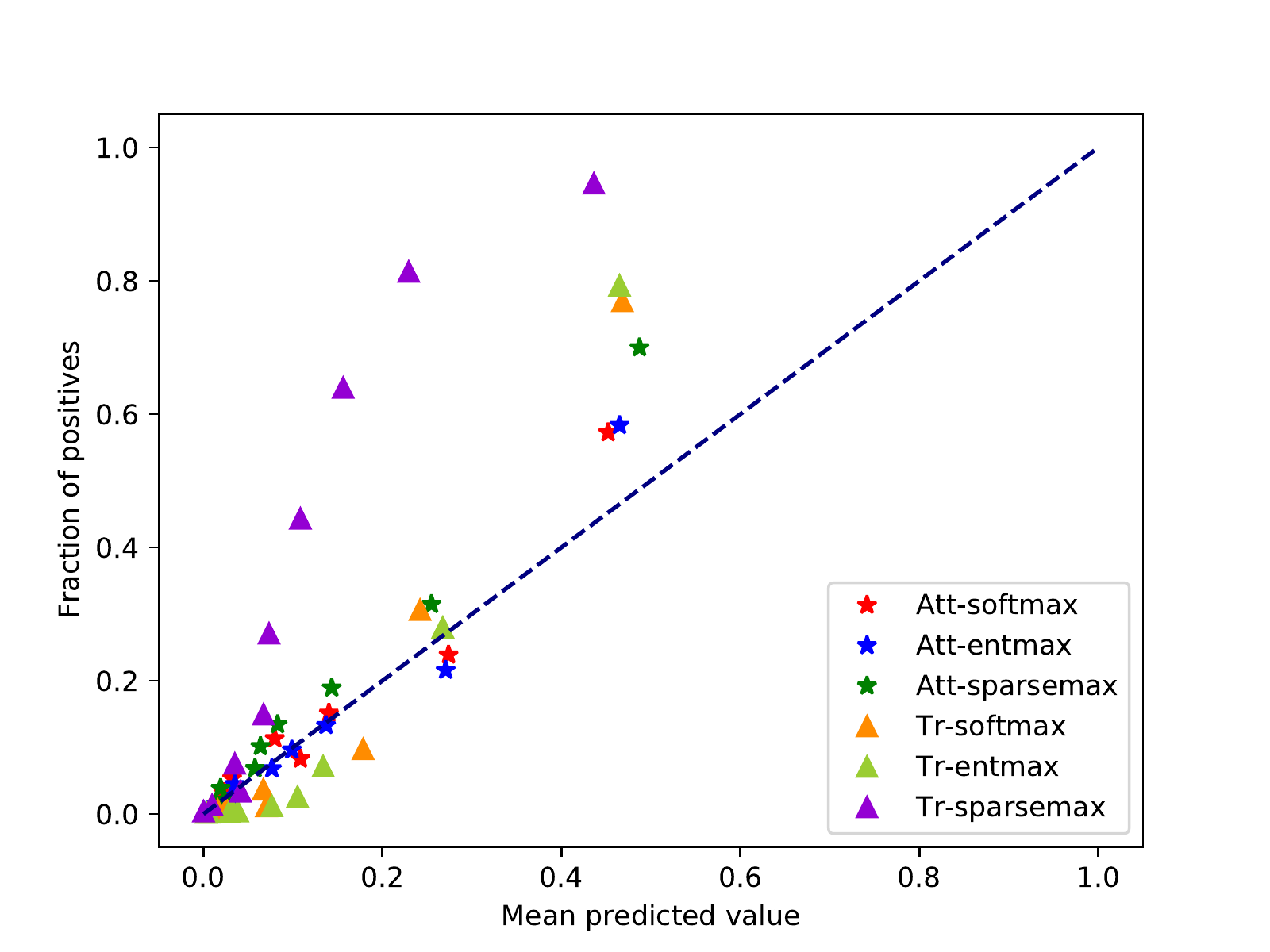} }}%
    \qquad
    
    \caption{Receiver operating characteristic curve (a) and calibration curve (b) for in-hospital mortality.}%
    \label{fig:inroc}%
\end{figure*}

\section{Experiments}

The  Medical Information Mart for Intensive Care  (MIMIC-III) database includes ICU information such as demographics, vital  measurements, laboratory test results, procedures, medications, caregiver notes, imaging reports, and mortality for critical care patients \citep{mimiciii}. \citet{Harutyunyan_2019} propose a public benchmark from MIMIC-III for modelling mortality, length of stay, physiologic decline, and phenotype classification. We use the in-hospital mortality benchmark cohort for extracting clinical notes in English based on the first 48 hours of an ICU stay to develop a prediction model. The cohort excludes ICU stays with missing length-of-stay, patients under 18, multiple ICU stays, stays under 48 hours, and without observations on the first 48 hours. The mortality class is defined by comparing the date of death across hospital admissions and discharge times with a resulting mortality rate of  13.2\%.
To preprocess the extracted notes, we tokenize with NLTK\footnote{NLTK tokenizer for English: \url{https://www.nltk.org/index.html}}, lowercase, and exclude duplicate and discharge notes. 
The training/validation/test datasets consist of data from $14,681$, $3,222$ and $3,236$ patients, respectively. We report the mean and standard deviation with 5 random runs for the area under the receiver operator characteristic curve (AUC-ROC), area under the precision-recall curve (AUC-PR), and the Brier score.

 We pretrain GloVe \citep{pennington2014glove} word embeddings on the benchmark training split with: 100D, minimum frequency of 5 words, window of 15 words, 20 epochs, and further fine-tune them on the downstream task.  For the prediction model, we use the following hyperparameters: Adam optimiser \citep{adamKingma}, learning rate $1\mathrm{e}{-4}$, epochs 30, hidden size 128, batch size 16, and dropout 0.2. For each training instance the max number of words $i$ in a sentence is 50 and the max number of sentences $t$ is 1000. Finally, we perform hyper-parameter and model selection  tuning with the validation dataset based on the AUC-ROC.

\begin{figure}[h]%
    \centering

    \subfloat[]{{\includegraphics[width=13.cm, height=13.cm,keepaspectratio]{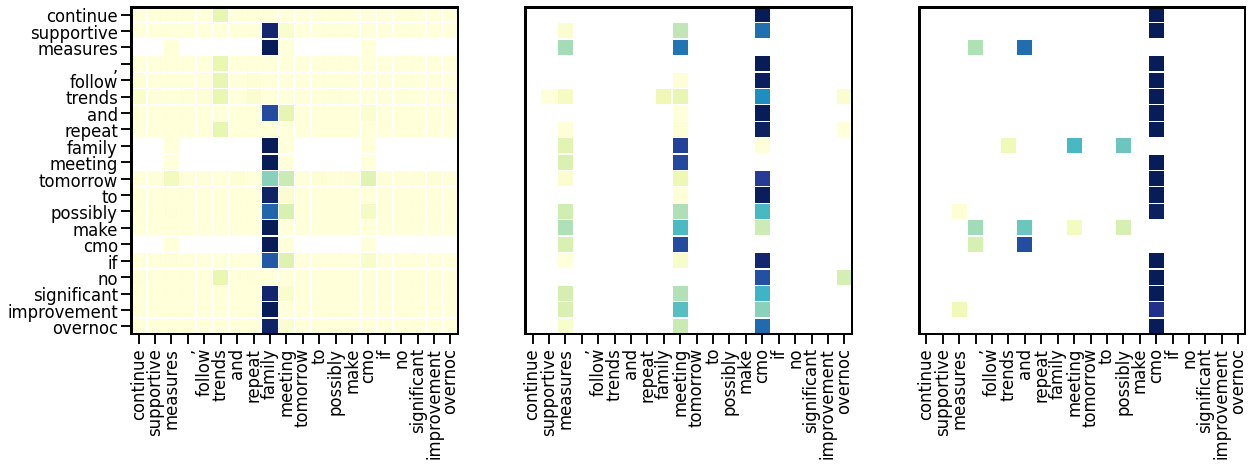} }}%
    \qquad
    \subfloat[]{{\includegraphics[width=13.cm, height=13.cm,keepaspectratio]{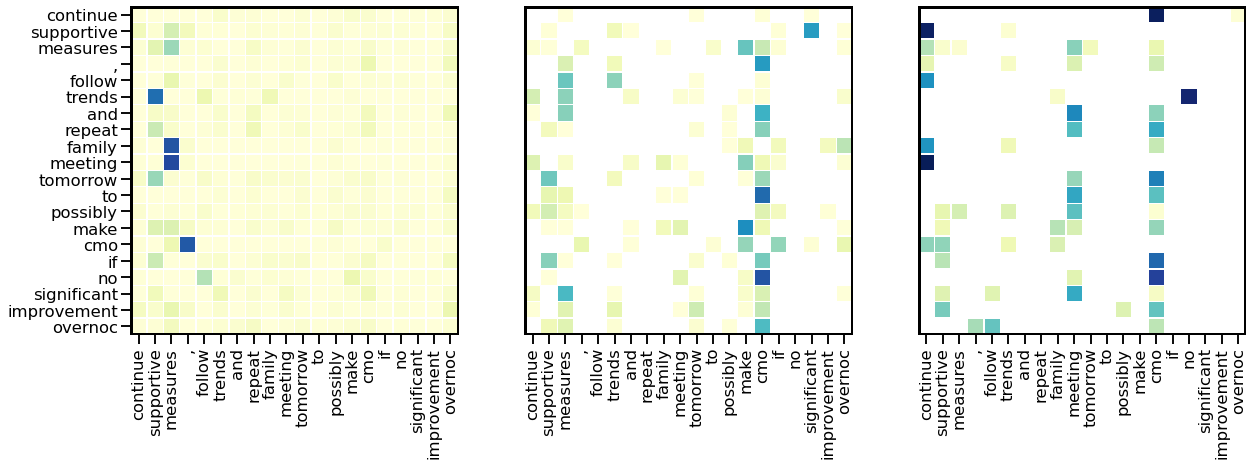} }}%
    
    \caption{Attention heatmaps corresponding  positive class instance. Att-softmax, Att-entmax and Att-sparsemax respectively in (a). Tr-softmax, Tr-entmax and Tr-sparsemax in (b).}%
    \label{fig:att_map1}%
\end{figure}

\subsection{In-hospital Mortality Results}

We report the performance on the task of in-hospital mortality prediction using sparse attention, and compare it to dense attention. Table \ref{Tab:inhospital} shows the AUC-ROC, AUC-PR, and Brier score results of the validation and test datasets. The Att-sparsemax shows competitive results compared to entmax and softmax on both metrics AUC-ROC and AUC-PR. The Brier score increases for the sparse attention models, where the models with sparse attention assign higher output probabilities to both classes. Figure \ref{fig:inroc} shows the ROC and calibration curves. The sparse attention models show competitive performance compared to the baseline on ROC. However, we observe from the calibration curve that the sparse attention underestimates predictions for the local attention models.

Moreover, the addition of sparse attention on the second level of the hierarchy for the transformer models, hurts the performance measures and calibration. The sparse transformer models show a marked miscalibration in Figure \ref{fig:inroc} where it underpredicts the true probabilities.

\subsection{Qualitative Analysis}

We show example attention heatmaps from the local self-attention model for sentences that include any of the following directives: do not resuscitate (dnr), do not intubate (dni), and comfort measures only (cmo). The directive words are known to be strongly correlated with mortality. For example, if the family of a patient has decided for comfort measures only then this refers to palliative treatment of a patient thought to be dying.

\citet{jain-wallace-2019-attention} define explainable attention when the inputs with the highest attention weights are relevant for a given prediction. 
Figure \ref{fig:att_map1} shows attention heatmaps for two sentences that contain any of the directive words.  The heatmaps denote the relative importance for each word embedding (row) with respect to its context (column), the darker the cell, the higher the weight.  One can observe that the sparse attention put most of its weights on the directive words, and that the highest attended words differ between the sparse and dense mechanisms. However, in Figure \ref{fig:att_map1} (a) $\entmax$ better distributes its weights for cmo. We show more attention examples in Appendix \ref{sect:attheatmap2}.
On the hierarchical transformer the sentence level sparsemax is consistently too aggressive in selecting weights. For example, in the sentence from Fig. \ref{fig:att_map1} (b) the model only assigns  attention to 8 out of 174 sentences.  
Moreover, the attention weights at the word encoder level shifted away from the directive words compared to the local attention.

\section{Discussion and Future Work}

In this paper, we compared the standard dense attention with sparse attention mechanisms on two setups of local and global dependencies across representations for mortality prediction with clinical notes. 
The sparsemax attention mechanism shows competitive predictive performance compared to the dense attention weights on the local self-attention model. However, the global model based on sparse attention produces under-confident predictions.  

The drop in performance in the hierarchical transformer is due to the sparse attention on the sentence encoder only assigning weights to few sentences, where most ICU stays contain hundreds of sentences.

Future work meriting investigation includes combining sparse and dense mechanisms on the hierarchical transformer. In addition to showing examples of the attention heatmaps, one may also analyse the predicted weights by measuring their correlation  to feature importance metrics \citep{vashishth2019attention, serrano-smith-2019-attention}.

\bibliography{references}
\bibliographystyle{acl_natbib}

\appendix


\section{Architectures}
\label{sect:apparch}

\begin{figure}[h]%
    \centering
     \includegraphics[width=4.7cm]{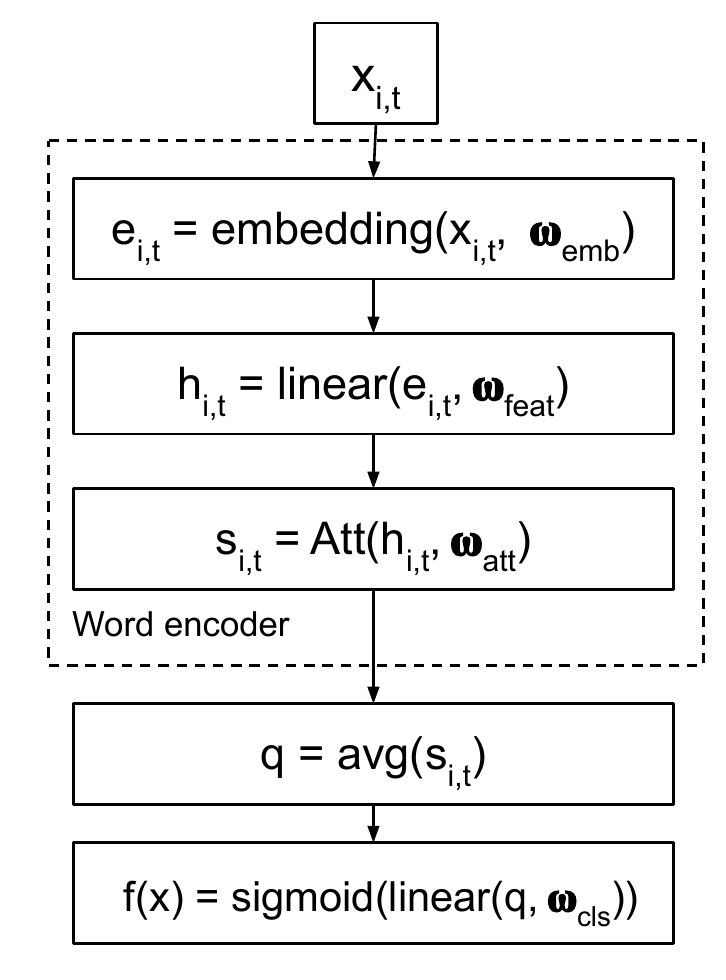} 
    
    \caption{Self-attention architecture.}%
    \label{fig:att_arch}%
\end{figure}

 We define the following architectures given an input $x_{i,t}$ that consist of clinical notes for an ICU stay represented into $i$ words and $t$ sentences.  
\paragraph{Self-attention} Figure \ref{fig:att_arch} shows the architecture for the self-attention model.

\begin{figure}[h]%
    \centering
     \includegraphics[width=4.7cm]{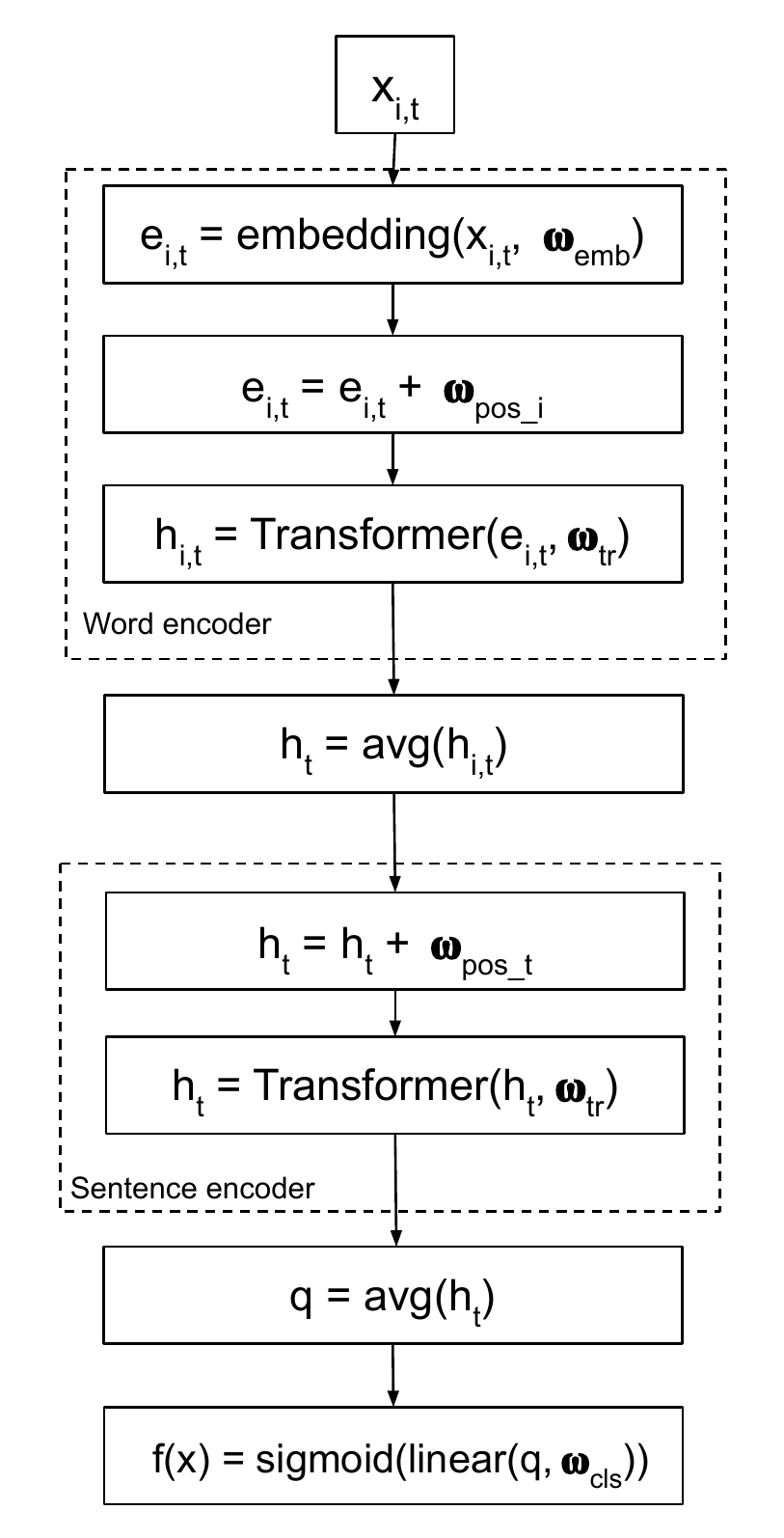} 
    
    \caption{Hierarchical transformer architecture.}%
    \label{fig:tr_arch}%
\end{figure}

\paragraph{Hierarchical Transformer} Figure \ref{fig:tr_arch} shows the architecture for the hierarchical transformer model. The architecture consists of pre-trained emebddings, position embeddings for words $\omega_\text{pos\_i}$, sentences $\omega_\text{pos\_t}$, and transformer layers with self-attention.

\section{Attention Heatmaps}
\label{sect:attheatmap2}
\begin{figure}[h]%
    \centering
     \subfloat[]{{\includegraphics[width=13.3cm, height=13cm,keepaspectratio]{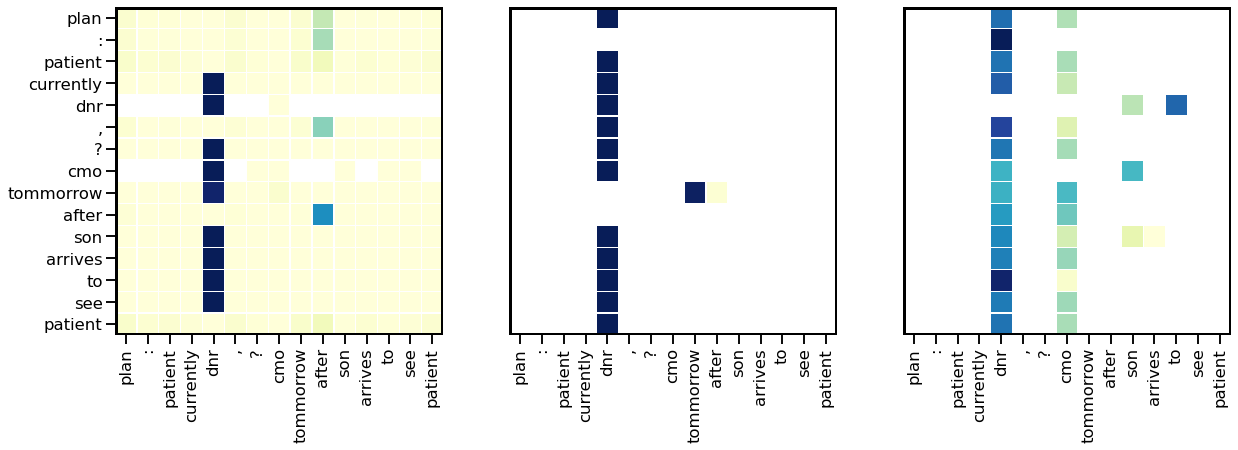} }}%
    
    \subfloat[]{{\includegraphics[width=13.3cm, height=13cm,keepaspectratio]{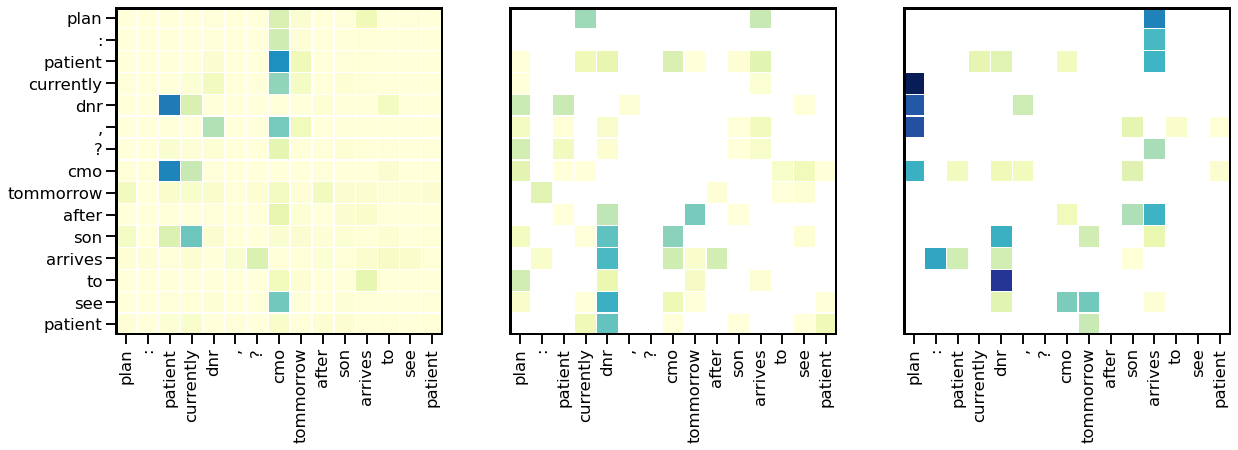} }}%
    \qquad
     \subfloat[]{{\includegraphics[width=13.3cm, height=13cm,keepaspectratio]{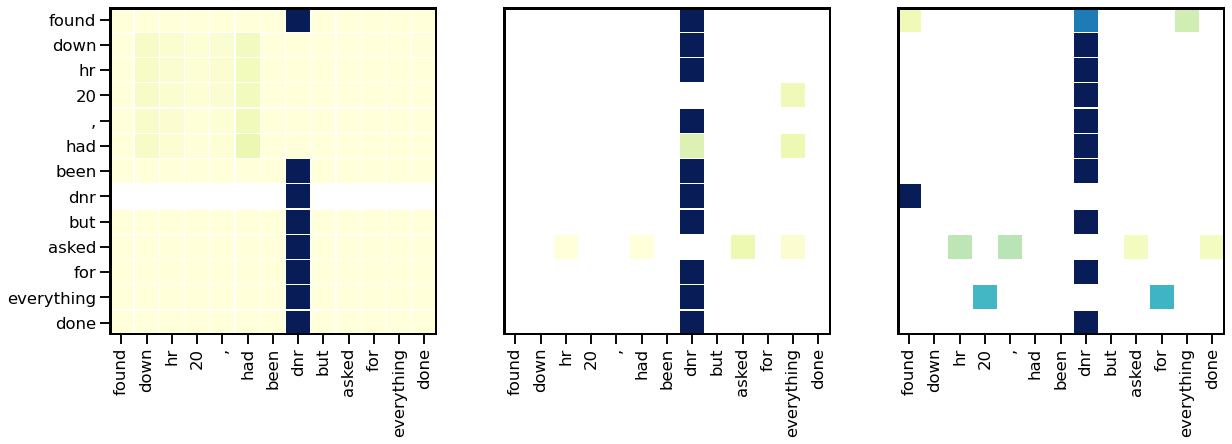} }}
     \qquad
     \subfloat[]{{\includegraphics[width=13.3cm, height=13cm,keepaspectratio]{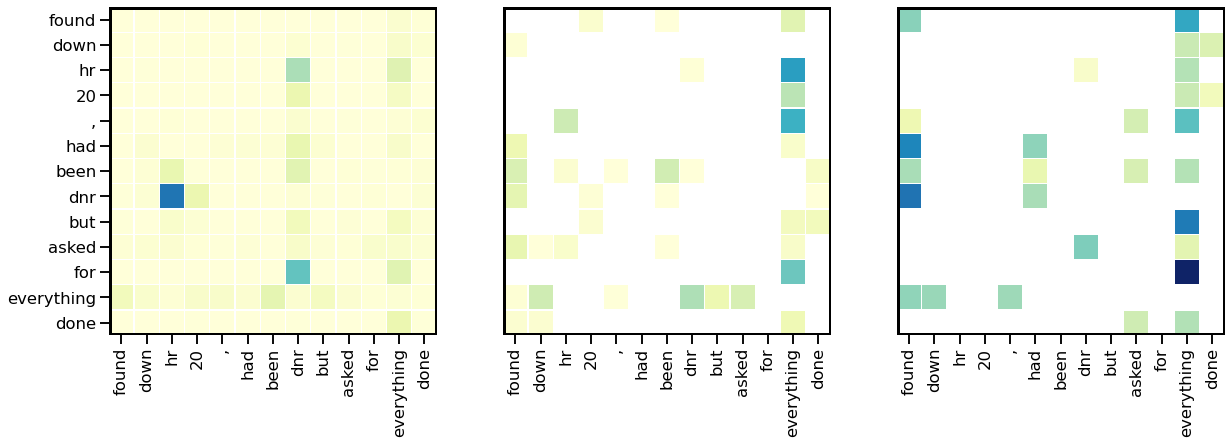} }}
    \caption{Attention heatmaps for sentences in a positive instance (a) Att-softmax, Att-entmax and Att-sparsemax respectively,  and (b) Tr-softmax, Tr-entmax and Tr-sparsemax. Sentences in negative instance (c) Att-softmax, Att-entmax and Att-sparsemax and (d) Tr-softmax, Tr-entmax and Tr-sparsemax .}%
    \label{fig:att_map2}%
\end{figure}

\end{document}

%% file: macros.tex
\DeclareMathOperator{\softmax}{softmax}
\DeclareMathOperator{\sparsemax}{sparsemax}
\DeclareMathOperator{\entmax}{entmax}

\DeclareMathOperator{\Att}{Att}